\pdfoutput=1
\documentclass[11pt]{article}

\usepackage{naacl2021}

\usepackage{times}
\usepackage{latexsym}
\usepackage{amsmath}
\usepackage{amssymb}
\usepackage{fancyvrb}
\usepackage{todonotes}
\usepackage{adjustbox}
\usepackage{multirow}
\usepackage{todonotes}

\usepackage{listings}

\definecolor{codegreen}{rgb}{0,0.6,0}
\definecolor{codegray}{rgb}{0.5,0.5,0.5}
\definecolor{codepurple}{rgb}{0.58,0,0.82}
\definecolor{backcolour}{rgb}{0.95,0.95,0.92}
\lstdefinestyle{mystyle}{
	backgroundcolor=\color{backcolour},   
	commentstyle=\color{codegreen},
	keywordstyle=\color{magenta},
	numberstyle=\tiny\color{codegray},
	stringstyle=\color{codepurple},
	basicstyle=\ttfamily\footnotesize,
	breakatwhitespace=false,         
	breaklines=true,                 
	captionpos=b,                    
	keepspaces=true,                 
	numbers=left,                    
	numbersep=5pt,                  
	showspaces=false,                
	showstringspaces=false,
	showtabs=false,                  
	tabsize=2,
	postbreak=\mbox{\textcolor{red}{$\hookrightarrow$}\space},
}
\usepackage[T1]{fontenc}
\usepackage[utf8]{inputenc}

\usepackage{microtype}

\title{MQDD: Pre-training of Multimodal Question Duplicity Detection for Software Engineering Domain}

\author{Jan Pašek, Jakub Sido, Miloslav Konopík, Ondřej Pražák\\[0.2em]
\tt{\{pasekj,sidoj,konopik,ondfa\}@kiv.zcu.cz}\\[0.5em]
 NTIS -- New Technologies for the Information Society, \\
Department of Computer Science and Engineering, \\
Faculty of Applied Sciences, University of West Bohemia, Plze\v{n}, Czech Republic
}

\begin{document}
\maketitle
\begin{abstract}
This work proposes a new pipeline for leveraging data collected on the Stack Overflow website for pre-training a multimodal model for searching duplicates on question answering websites. Our multimodal model is trained on question descriptions and source codes in multiple programming languages. We design two new learning objectives to improve duplicate detection capabilities. The result of this work is a mature, fine-tuned Multimodal Question Duplicity Detection (MQDD) model, ready to be integrated into a Stack Overflow search system, where it can help users find answers for already answered questions. Alongside the MQDD model, we release two datasets related to the software engineering domain. The first Stack Overflow Dataset (SOD) represents a massive corpus of paired questions and answers. The second Stack Overflow Duplicity Dataset (SODD) contains data for training duplicate detection models.
\end{abstract}

\section{Introduction}\label{sec:introduction}

The benefits of Question-Answer (QA) networks for software developers such as the Stack Overflow website are widely exploited by professionals and beginners alike during the software creation process. Many solutions to various problems, short tutorials, and other helpful tips can be found on these networks. However, access to this valuable source of information highly depends on users' ability to search for the answers. In our paper, we introduce a multimodal method for detecting duplicate questions. Apart from the primary use to prevent posting duplicate questions, this technique can be directly used for better search. When users are posting an already answered question, they can get the answer immediately without the necessity to wait until someone else links the duplicate post or answers their question.

The duplicate question detection task aims to classify whether two questions share the same intent. In other words, if two questions are duplicates, they relate to the same answer. The duplicate detection task is quite challenging since the classifier needs to distinguish tiny semantic nuances that can significantly change the desired answer.

The posts in the QA networks for software development mix natural language and source code snippets. The great success of neural networks for Natural Language Processing (NLP) encourages us to employ these methods for our task where natural language is intermixed with source codes. We support our decision with the findings of \citet{src:on-naturalness}, who show that the source codes carry even less entropy than the English language.
% We argue that the NLP methods should be capable of processing source codes 

Current state-of-the-art NLP methods build on large pre-trained models based on the Transformer architecture \cite{src:transformer}. The Transformer-based models such as BERT \cite{src:bert}, GPT \cite{src:GPT}, RoBERTa \cite{src:roberta}, or T5 \cite{src:T5} are usually pre-trained on massive unlabeled corpora and applied to a task with much less training data afterward. We follow this idea and introduce the pre-training phase into our solution.

In our work, we try to leverage the transfer learning techniques to build a duplicate question detection \cite{src:duplicate-detection-lstm} model for question-answering platforms such as Stack Overflow\footnote{\url{https://stackoverflow.com}} or Quora\footnote{\url{https://quora.com/}}. Since both platforms are extensively used to ask programming-related questions, we aim to pre-train a multimodal encoder for a programming language (PL) and natural language (NL). To achieve the best possible results, we design duplicate-detection-specific pre-training objectives (see Section \ref{sec:pretrainig-objectives}). % These are \textit{Question-Answer} (QA) and \textit{Same Post} (SP). These tasks require the model to fully understand the provided multimodal input and force it to distinguish tiny semantic nuances. In addition to those specific objectives, we employ the \textit{Masked Language Modeling} \cite{src:bert} utilized by numerous related works. 

Since the source code snippets present in the Stack Overflow questions may be relatively long, we choose to base our model on the Longformer architecture \cite{src:longformer}; whose modified attention scheme scales linearly with the sequence length. The resulting model with $\approx$146M parameters is firstly pre-trained on a large semi-supervised corpus of Stack Overflow questions and answers. For detailed information about the dataset and pre-training, see Section \ref{sec:pretraining}. 

\begin{figure*}[htbp]
\centering
\includegraphics[width=0.9\linewidth]{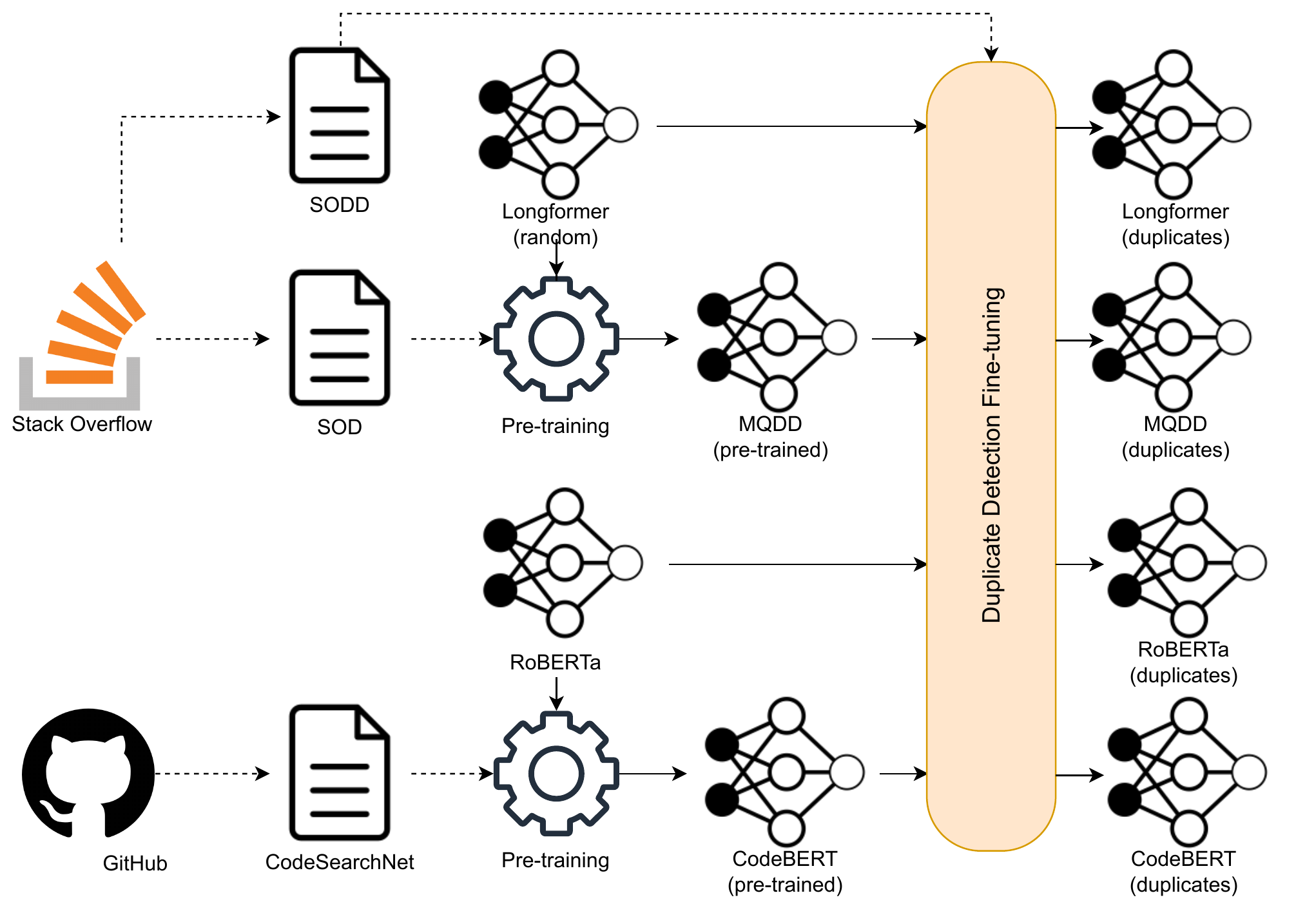}
\caption{A visualization of the pipeline of our experiments. The upper part of the figure shows the construction of our SOD and SODD datasets and their usage for pre-training and fine-tuning our MQDD model. The lower part of the figure visualizes the pre-training of the CodeBERT done by \citet{src:code_bert_orig}.}
\label{fig:system-overview}
\end{figure*}

Afterward, in Section \ref{sec:duplicates}, we fine-tune the obtained model on the duplicate detection task and compare our model with CodeBERT \cite{src:code_bert_orig}, which represents another NL-PL multimodal encoder. We also compare our model to a randomly initialized Longformer \cite{src:longformer} and pre-trained RoBERTa \cite{src:roberta} to see whether the pre-training of both models brings a significant improvement of the achieved results. The previously described experiments are visualized in Figure \ref{fig:system-overview}. At the end of this paper, we explore how well our model generalizes to other tasks by applying our model to the CodeSearchNet dataset \cite{src:code-search-net} in Section \ref{sec:generalization}.

%%% --> abstract? 

Our main contributions are: 1) We release a fine-tuned Multimodal Question Duplicity Detection (MQDD) model for duplicate question detection. The model is mature enough to be deployed to Stack Overflow, where it can automatically link duplicate questions and, therefore, improve users' ability to search for desired answers. Furthermore, we release the pre-trained version of the encoder, so other researchers may reuse the most computationally intensive phase of our model training. 2) We present and explore the effect of entirely new pre-training objectives specially designed for duplicate detection. 3) We release a \textit{Stack Overflow Dataset} (SOD) that can be used for pre-training models in a software engineering domain. Furthermore, we release a novel \textit{Stack Overflow Duplicity Dataset} (SODD) for duplicate question detection, enabling other researchers to follow up on our work seamlessly.

\section{Related Work}\label{sec:related-work}
Deep-learning models have become more common in many new domains in recent years. The main advantage lies in processing a large volume of documents and benefiting from them. Software development is not an exception. The naturally collected massive amount of data in software management systems, issue tracker tools, and versioning systems makes the software development an ideal domain to apply deep models to increase work effectiveness.
% \paragraph{SWI}

Developers from Microsoft have introduced a system called DeepCodeReviewer, which recommends the proper review from a repository of historical peer reviews from their database \cite{gupta2018intelligent}. 
\citet{ullah2019source} have outperformed the existing techniques in source code authorship attribution using deep learning methods to express coding styles’ features for each programmer and attribute the actual authors.
\citet{tufano2018deep} use learned vector embedding to create vector representations of source codes to detect similarities in code. 
\citet{yin2017syntactic, balog2017deepcoder} utilize RNN for source code generation from natural language.
Authors of Commit2Vec \cite{cabrera2021commit2vec} use analysis of code structure before and after repository commit to reveal a change in the security of the source code. \citet{liu2020atom} generate commit messages from code changes. \citet{deng2020structure} experiment with SQL from natural language generation. Another representative of source code-oriented models is the Codex \cite{src:codex}, a massive pre-trained model designated for source code generation. Its slightly modified form is also integrated with the \textit{GitHub Copilot}\footnote{\url{https://copilot.github.com}} system, a digital pair programmer. \citet{src:code-t5} introduces a generative model called CodeT5, capable of solving multiple tasks thanks to converting all problems into a unified sequence-to-sequence form.
\citet{sun2022code} translate source codes into a natural language to retrieve similar code snippets.

% Code search
% \cite{gu2018deep} Deep code search 

% Deep Learning Similarities from Different Representations of Source Code  2018 https://ieeexplore.ieee.org/stamp/stamp.jsp?tp=&arnumber=8595238

% Automated Vulnerability Detection in Source Code Using Deep Representation Learning 2018 https://ieeexplore.ieee.org/stamp/stamp.jsp?tp=&arnumber=8614145

% Mention \cite{src:code_bert_orig, src:source_code_embedding_literature, src:longformer, src:transformer, src:bert}
 
% Comparison of Image-Based and Text-Based Source Code Classification Using Deep Learning 2020 https://link.springer.com/article/10.1007/s42979-020-00281-1
% sračka -- kralovina  málo ci tací 3 vydání 

% Deep learning code fragments for code clone detection 2016  https://ieeexplore.ieee.org/abstract/document/7582748

In NLP, many recent successful models follow the BERT \cite{src:bert} architecture and use the Transformer encoder \cite{src:transformer} to produce contextual representations of input tokens. These contextual embeddings \cite{src:contextual-1, src:contextual-2} can then be used for various tasks, including the classification of entire sequences \cite{src:bert-textual-similarity} or individual tokens \cite{src:bert-ner,src:bert-sentiment}. Such success can probably be attributed to a well-used attention mechanism \cite{src:attention-mechanism}, which allows the model to capture contextual information from the entire sequence being processed.

% Examples of possible applications are named entity recognition \cite{src:bert-ner}, sentiment analysis \cite{src:bert-sentiment}, the semantic similarity of sentences \cite{src:bert-textual-similarity}, etc. 
% \paragraph{objective}
In deep learning, the correct choice of the training objective significantly impacts the accuracy of a final system. Adapting the pre-training phase and finding a proper objective allows the model to exploit useful features from large, naturally labeled data enabling future gains on a downstream task. An example of a model that tries to improve the pre-training objectives is ELECTRA \cite{src:electra}, which employs a generator-discriminator setup. RoBERTa \cite{src:roberta} slightly modifies the Masked Language Modeling (MLM) objective and abandons the Next Sentence Prediction (NSP). Further work then tries to modify the used neural network architecture and thus target various problems. For example, Longformer \cite{src:longformer} and Reformer \cite{src:reformer} models significantly modify the attention mechanism to mitigate the $\mathcal{O}(N^2)$ complexity of a vanilla attention.

With increasing computational capability and model complexity, we can observe more multilingual pre-training goals to achieve better results --
% merging models then seek to introduce a variety of enhancements that allow the processing of multilingual or very long sentences, or they aim to improve pre-training goals to help achieve better results. 
% \paragraph{multiling and bimodal}
mBERT \cite{src:bert} and SlavicBERT \cite{src:slavic-bert} create a joint model for representing several different languages.
% \todo{multi-modal or multi-moda? :-) Za mě multi-modal OK (Honza)}
%multilingual models, Bi-modal models are very close to multilingual models. These, unlike
Unlike multilingual models, multimodal models process inputs from entirely different domains. For example, VilBERT \cite{src:vilbert} and VideoBERT \cite{src:videobert} process inputs consisting of both the natural language and video.

Another possible usage of multimodal encoders is to process inputs consisting of source codes and natural texts. Such models can produce contextual embeddings of source codes \cite{src:source_code_embedding_literature} directly applicable to downstream tasks such as programming language detection, code similarity, code search, or code fixing \cite{src:deep-learning-source-code}. Furthermore, such a multimodal encoder can be incorporated in a Seq2Seq setup, in which it can, for example, generate source code or its documentation.

The CuBERT \cite{src:cubert} is an example of a multimodal encoder for Python source codes and texts. It is pre-trained on 7.4M source files. The model outperforms BiLSTM \cite{src:bi-rnn, src:LSTM} and randomly initialized Transformer \cite{src:transformer} approach in five different tasks, including classification of variable misuse, wrong binary operator usage, swapped operands, and function-docstring match. Another representative of multimodal source code encoders is the CodeBERT model \cite{src:code_bert_orig} pre-trained on a multilingual corpus of source codes from six different programming languages. The CodeBERT builds upon the RoBERTa \cite{src:roberta} and follows the generator-discriminator approach laid out in ELECTRA \cite{src:electra}. Besides the replaced token detection (RTD) learning objective from ELECTRA, the authors employ the traditional MLM objective. The resulting model shows superior results in code search, natural language-programming language (NL-PL) probing, and documentation generation.

Our work differs from the previous multimodal source code encoders in the following points: 1) Our model is trained using novel pre-training objectives targeting specifically the duplicate detection task. 2) Unlike the CuBERT, explicitly designated for Python and CodeBERT, pre-trained on six different programming languages, our model is capable of processing inputs from an arbitrary programming language. Processing every possible programming language is crucial for deploying such a model to real-world question-answering platforms. 3) Our MQDD model employs a Transformer-based architecture with an attention scheme scaling linearly with sequence length. This enables the model to be integrated into systems that require processing long sequences in a reasonable time.

\section{Model Pre-training}\label{sec:pretraining}
This section describes the pre-training procedure, including the construction of the new dataset from the Stack Overflow, the definition of the learning objectives, and the model itself. 

\subsection{Stack Overflow Dataset}\label{sec:pretraining-dataset}
For the pre-training, we construct our Stack Overflow Dataset (SOD), created from the Stack Overflow data dump\footnote{Available at: \url{https://archive.org/download/stackexchange}.}. The original data source\footnote{Data dump was downloaded in June 2020. Therefore, all the stated information is valid to this date.} contains around 45M of posts (questions + answers) exported in an XML format. The questions represent approximately 17.7M thereof. To construct the dataset, we take all question-answer pairs, extract the textual and source code parts and apply different pre-processing on both (for pre-processing details, see appendix \ref{sec:dataset-preprocessing}). A result of the pre-processing procedure are \textit{tuples} $(Q_t, Q_c, A_t, A_c)$ containing pre-processed texts ($t$) and codes ($c$) from both the questions ($Q$) and answers ($A$).

\begin{table}[h!]
\begin{center}
\begin{tabular}{ccr}
\hline
\textbf{Order} & \textbf{Tag}   & \multicolumn{1}{c}{\textbf{Percentage}} \\ \hline
1              & javascript     & 10,95                                   \\
2              & java           & 9,88                                    \\
3              & c\#            & 8,04                                    \\
4              & php            & 7,95                                    \\
5              & python         & 6,32                                    \\
6              & html           & 6,18                                    \\
7              & css            & 4,28                                    \\
8              & c++            & 4,15                                    \\
9              & sql            & 3,42                                    \\
10             & c              & 2,29                                    \\
11             & objective-c    & 1,93                                    \\
12             & r              & 1,36                                    \\
13             & ruby           & 1,32                                    \\
14             & swift          & 1,05                                    \\
15             & bash           & 0,8                                     \\ \hline
-              & \textit{total} & 69,92                                   \\ \hline
\end{tabular}
\caption{The table presents a tag-based analysis of the percentage of individual programming languages in the SOD dataset. The table shows the 15 most frequent programming languages included in the dataset. Together they form $\approx$70\% of all the examples. The remaining 30\% are then made up of less popular programming languages or specific technologies.}
\label{tab:dataset-language-analysis}
\end{center}
\end{table}

Afterward, we construct the training set by taking \textit{2-combinations} of the pre-processed \textit{tuples}, resulting in 6 different \textit{input pair} types described in Section \ref{sec:pretrainig-objectives}. The acquired \textit{input pairs} ($x_1, x_2$) are further processed in batches of $100$ examples. For each pair in the batch, we sample one negative example by choosing a random text or code $x_r$ from the batch buffer and use it as a replacement for the second element in the pair. This results in adding pair ($x_1, x_r$) to the training set.

Subsequently, we tokenize the input pairs and store them in \textit{TFRecords}\footnote{\url{https://www.tensorflow.org/tutorials/load_data/tfrecord}} to speed up data processing during the training. The resulting dataset contains 218.5M examples and can be downloaded from our GitHub repository \url{https://github.com/kiv-air/StackOverflowDataset}. A detailed description of the dataset's structure is provided in appendix \ref{sec:sod-structure}. Furthermore, Tables \ref{tab:dataset-language-analysis} and \ref{tab:sod-dataset-statistics} present a detailed analysis of the programming languages included in the corpus and dataset size, respectively.

\begin{table*}[]
\catcode`\-=12
\begin{adjustbox}{width=0.57\linewidth,center}
\begin{tabular}{lrrrrc}
\hline
\multicolumn{1}{c}{\textbf{Statistic}} &
  \multicolumn{1}{c}{\textbf{QC}} &
  \multicolumn{1}{c}{\textbf{QT}} &
  \multicolumn{1}{c}{\textbf{AC}} &
  \multicolumn{1}{c}{\textbf{AT}} &
  \textbf{Total} \\ \hline
average \# of characters & 846 & 519 & 396 & 369 & -                       \\
average \# of tokens     & 298 & 130 & 140  & 92  & -                       \\
average \# of words      & 83  & 89  & 44  & 60  & -                       \\
\# of characters         & 16.1B & 13.5B & 6.6B & 9.6B & \multicolumn{1}{r}{45.8B} \\
\# of tokens             & 5.7B & 3.4B & 2.3B & 2.4B & \multicolumn{1}{r}{13.8B} \\
\# of words              & 1.6B & 2.3B & 0.7B & 1.6B & \multicolumn{1}{r}{6.2B} \\ \hline
\end{tabular}
\end{adjustbox}
\caption{Detailed statistics of the released Stack Overflow Dataset (SOD). The table shows the average number of characters, tokens, and words in different source codes present in questions (QC) or answers (AC) and texts present in questions (QT) or answers (AT). Besides the average statistics, the table provides a total count of tokens, words, or characters. To calculate the statistics related to token counts, we utilized the tokenizer presented in Section \ref{sec:pretraining-tokenization}, whereas we employed a simple space tokenization for the word statistics.}
\label{tab:sod-dataset-statistics}
\end{table*}

\subsection{Tokenization}\label{sec:pretraining-tokenization}
Before extracting the input pairs, we employ the $(Q_t, Q_c, A_t, A_c)$ tuples to train a joint tokenizer for both the source codes and English texts. We use the \textit{Word Piece} tokenizer \cite{src:word-piece}, whose vocabulary size is typically set to a value between 10K-100K subword tokens. In our work, we set the vocabulary size to 50K subword tokens, which is large enough to encompass both the textual and code tokens while preserving a reasonable size of the embedding layer. When constructing the dataset, we ignore all tokens that occur less than five times in the dataset.

\subsection{Pre-training Objectives}\label{sec:pretrainig-objectives}
Similarly to BERT \cite{src:bert}, we employ a \textit{Masked Language Modeling (MLM)} task during the pre-training phase. The \textit{MLM} objective aims to reconstruct original tokens from intentionally modified input sequences. The modification replaces randomly selected tokens with a special \verb+[MASK]+ token or any other token from the dictionary.

Besides the \textit{MLM}, we introduce two Stack Overflow dataset-specific tasks dealing with multimodal data. The first task is called \textit{Question-Answer (QA)}, and it aims is to classify whether the \textit{input pair} originates from a question-answer relationship. The individual elements of the \textit{input pair} can be either a natural language text or a programming language snippet. Therefore, we work with the following \textit{input pair} types:

        \begin{itemize}
            \item Question text - Answer code \textit{(Qt-Ac)}
            \item Question code - Answer code \textit{(Qc-Ac)}
            \item Question text - Answer code \textit{(Qt-Ac)}
            \item Question code - Answer text \textit{(Qc-At)}
        \end{itemize}

The second Stack Overflow-related task is called \textit{Same Post (SP)}. Similarly to the \textit{QA} task, the \textit{SP} works with \textit{input pairs} of natural language and source code snippets. However, unlike the \textit{QA} task, \textit{SP} classifies whether the elements of the \textit{input pair} come from the same post (a post represents either a question or an answer). The resulting possible \textit{input pair} types are the following:

        \begin{itemize}
            \item Answer Text - Answer Code \textit{(At-Ac)}
            \item Question Text - Question Code \textit{(Qt-Qc)}
        \end{itemize}

We designed these learning objectives specifically to achieve the best possible result on our target task - \textit{duplicate detection} (Section \ref{sec:duplicates}). We presume that employing these tasks requiring a deep understanding of the multimodal input helps us outperform similar models such as CodeBERT \cite{src:code_bert_orig}. Furthermore, our learning objectives require comparing and matching the semantics of both the textual input and the source code that can be leveraged on downstream tasks such as \textit{code search} \cite{src:code-search-1, src:code-search-2, src:code-search-3}.

\subsection{Model Description}\label{sec:pretraining-model-description}
We choose to employ the architecture of the \textit{Longformer} model \cite{src:longformer} for its attention mechanism that scales linearly with the input sequence length. This addresses the fact that the processed input sequences (mainly the source code) may contain several hundreds of tokens. Processing such long sequences with the vanilla attention mechanism used in the Transformer \cite{src:transformer} can be computationally exhausting. 

We use the \textit{Hugging Face's Transformers} \cite{src:huggingface} model with approximately 146M parameters (for more details on the model, see appendix \ref{sec:longformer-configuration}).

On top of the base model, we build two different classification heads. The first head, dealing with the \textit{MLM} task, takes the input tokens' contextual embeddings as its input. It means the \textit{MLM head} works with the matrix $\mathbf{E} \in \mathbb{R}^{N \times H}$, where $H$ is the hidden size and $N$ is the length of the input sequence. \textit{MLM} prediction is obtained by passing the matrix through a \textit{linear layer} so that $\mathbf{MLM_{output}} = E \times W_{mlm}$, where $W_{mlm} \in \mathbb{R}^{H \times |V|}$, and $|V|$ represents the size of the vocabulary. In other words, the model produces a probability distribution over the vocabulary for each of the input tokens, including the masked ones. To optimize the weights, we further calculate a cross-entropy loss over the network's prediction.

The second head classifies whether an input pair represents a \textit{question-answer pair} and whether both inputs originate from the \textit{same post}. To achieve this, the head takes the contextual embedding of the special \verb+[CLS]+ token (\verb+[CLS]+ $\in \mathbb{R}^H$)\footnote{The [CLS] token is an artificial token added at the begging for sequence classification tasks.}. The vector is then transformed using a \textit{linear layer} with \textit{ReLu} \cite{src:relu} used as an activation function - $\mathbf{QA\_SP_{intermediate}} = relu(\verb+[CLS]+ \times W_{qa\_sp_1})$, where $W_{qa\_sp_1} \in \mathbb{R}^{H \times D}$ and $D$ represents a dimensionality of the intermediate layer. In the end, the \textit{Question-Answer/Same Paragraph} (QA/SP) head output is obtained using another \textit{linear layer} - $\mathbf{QA\_SP_{output}} = QA\_SP_{intermediate} \times W_{qa\_sp_2}$, where $W_{qa\_sp_2} \in \mathbb{R}^{D \times 2}$. Put differently, the \textit{QA/SP} head is a multi-label classifier with two output neurons. The first one represents a probability of the input pair originating from the same post. The second one represents the probability of the input pair originating from the \textit{question-answer} relationship. To optimize the weights with respect to our \textit{QA/SP} objectives, we compute a binary cross-entropy loss over the two output neurons.

\subsection{Pre-training Procedure}\label{sec:pre-training-procedure}

We optimize our model using Adam optimizer \cite{src:adam} with a \textit{learning rate} of $1\mathrm{e}{-5}$ while employing both \textit{linear warmup} and \textit{linear decay} to zero. The \textit{linear warmup} is configured to reach the target \textit{learning rate} in 45K batches. The pre-training is carried out on two Nvidia A100 GPUs and two AMD EPYC 7662 CPU cores with a batch size of 64 examples.

We perform a single iteration over the whole dataset ($\approx220M$ examples) with such a configuration while trimming the sequences to a \textit{sequence length} of 256 tokens. Afterward, we set the \textit{sequence length} to 1024 tokens and train the model for additional 10M examples, enabling us to train positional embeddings for longer sequences.

\section{Duplicate Question Detection}\label{sec:duplicates}

Following the pre-training phase, this section focuses on applying the obtained model to the task of duplicate detection. In the first part, we describe the construction of a new dataset for duplicate detection. The next part presents how we integrate the pre-trained model into a two-tower neural network. At the end of this section, we describe the concluded experiments and present the results.

\subsection{Stack Overflow Duplicity Dataset}
Similarly to the pre-training phase, we employ the Stack Overflow data dump to assemble the Stack Overflow Duplicity Dataset (SODD). The data contain approximately 491K pairs of questions marked to be duplicated by the page's users. To replenish the dataset with negative samples, we employ randomly chosen questions and similar questions retrieved using ElasticSearch\footnote{\url{https://www.elastic.co}}. More specifically, we sample three random questions and retrieve six similar questions for each duplicate pair. The similarities are retrieved either based on a full-text similarity of the question's body or associated tags.  However, each question can be included in the dataset at most once. The resulting dataset consists of approximately 1.4M examples represented by triplets $(x_1, x_2, y)$, where $x_1$ and $x_2$ represent the questions and $y \in \{\textrm{\textit{duplicate, text\_similar, tag\_similar, different}}\}$ represents the label. Although the dataset differentiates between different and similar questions, all of our experiments treat both of these classes the same. In other words, our experiments perform a binary classification task. For more information about the dataset size, see Table \ref{tab:dataset_final_counts}.

\begin{table}[h!]
	\begin{center}
		\begin{tabular}{l || r r r r} 
			\hline
			\textbf{Type} & \textbf{Train} & \textbf{Dev} & \textbf{Test} & \textbf{Total} \\ [0.5ex] 
			\hline\hline
			Different & 550K & 64K & 32K & 646K \\
			Similar & 526K & 62K & 30K & 618K \\
			Duplicates & 191K & 22K & 11K & 224K \\
			\hline
			Total & 1.2M & 148K & 73K & 1.4M \\
			\hline
		\end{tabular}
	\end{center}
	\caption{Stack Overflow Duplicity Dataset (SODD) size summary.}
	\label{tab:dataset_final_counts}
\end{table}

The question pairs acquired from the Stack Overflow are stored in the \textit{HTML} format. Therefore, we employ a \verb|BeautifulSoup|\footnote{\url{https://beautiful-soup-4.readthedocs.io/en/latest/}} library to remove unwanted \textit{HTML} markup and separate normal text from source code snippets. Besides, we pre-process the source code stripping all inline comments and newline characters. Moreover, we substitute the floating-point and integer numbers with placeholder tokens. Similarly to the source codes, we replace numbers and date/time information with placeholder tokens and remove newlines and punctuation in the textual part of the dataset. The resulting dataset can be obtained from our repository \url{https://github.com/kiv-air/StackOverflowDataset}. For a detailed description of the dataset structure, see appendix \ref{sec:sodd-structure}.

\subsection{Model}

\begin{figure}[htbp]
\centering
\includegraphics[width=\linewidth]{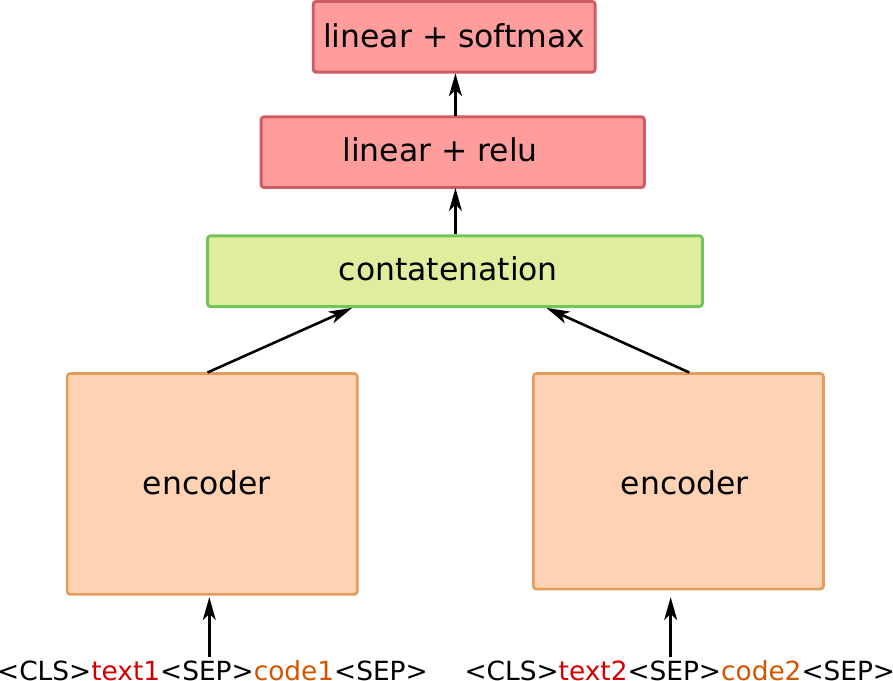}
\caption{The neural network model architecture used for duplicate question detection. The encoder blocks in the figure share the same weights and represent either an MQDD, CodeBERT \cite{src:code_bert_orig}, or RoBERTa \cite{src:roberta}.}
\label{fig:duplicate-architecture}
\end{figure}

We employ a variant of a two-tower neural network to adapt our pre-trained model to the duplicate detection task. Our setup (Figure \ref{fig:duplicate-architecture}) encodes both questions separately using the same pre-trained encoder, obtaining representations of the questions ($x_{e1}, x_{e2} \in \mathbb{R}^{d}$). The representations are then concatenated ($x_e = [x_{e1};x_{e2}]$) and transformed using a linear layer with ReLu activation \cite{src:relu}, as stated in equation \ref{eq:first-linear}.

\begin{equation}\label{eq:first-linear}
    x_L = max(0, x_e W_L + b_L)
\end{equation}

\begin{equation}\label{eq:cls-head}
    x_H = softmax(x_L W_H + b_H)
\end{equation}

At the top of our duplicate detection model, there is a classification head consisting of a linear layer with two neurons, whose activation is further transformed using a softmax function, \cite{src:softmax} as shown in equation \ref{eq:cls-head}.

An alternative approach would be to jointly pass both questions into the encoder and build a classification head at the top. However, our architecture of the two-tower model allows the representations of the whole corpus to be pre-computed and indexed in a fast vector space search library such as Faiss\footnote{\url{https://github.com/facebookresearch/faiss}} \cite{src:faiss} (see the future work in Section \ref{sec:future_work}). Thanks to that, it is possible to compute only the representation of the newly posted question and run a quick search inside the vector space. This is much faster than running the model for each pair of questions composed of a new question and the others in the corpus.

\subsection{Experimental Setup - Duplicate Detection}
Similarly to the pre-training phase, we use the Adam optimizer \cite{src:adam} with a learning rate set to $6.35\mathrm{e}{-6}$ to train the model on a computation node with two cores of AMD EPYC 7662 CPU and two Nvidia A100 GPUs. In each experiment, we train the model for 24 hours with a batch size of 96 examples and observe the progress of cross-entropy loss, accuracy, and F1 score. The hyperparameters were set based on 30 hyperparameter-search experiments conducted using the \textit{Weights \& Biases} \cite{src:wandb} \textit{sweeps} service\footnote{\url{https://docs.wandb.ai/guides/sweeps}}. For detailed information about the hyperparameter setting, refer to appendix \ref{sec:duplicates-hyperparameters}.

To evaluate the effectiveness of our pre-training objectives, we compare our model with the \textit{CodeBERT} \cite{src:code_bert_orig}, RoBERTa \cite{src:roberta}, and randomly initialized Longformer \cite{src:longformer}. The comparison experiments also utilize the architecture depicted in Figure \ref{fig:duplicate-architecture}, where we only replace the encoder with the model being compared. The training setup for the comparison experiment is identical to the setup described above. It means that we fine-tune the models for 24 hours on the same hardware.

\subsection{Results}\label{sec:duplicates-results}
As evaluation metrics, we use an \textit{F1 score} and \textit{accuracy}. We summarize the results of our experiments in Table \ref{tab:duplicates-results}, where the achieved results are stated with 95\% confidence intervals. From the results, we can see that our model significantly outperformed all alternative approaches. For further discussion on the results, see Section \ref{sec:discussion}.

\begin{table}[ht!]
\begin{tabular}{lrr}
\hline
\multicolumn{1}{l}{\textbf{Model}} & \multicolumn{1}{c}{\textbf{Accuracy}} & \multicolumn{1}{c}{\textbf{F1 Score}} \\ \hline
\textbf{MQDD}                              & \textbf{74.83} $\pm$ \textbf{0.10}                     & \textbf{75.10} $\pm$ \textbf{0.10}                    \\
CodeBERT                           & 70.44 $\pm$ 0.12                     & 70.70 $\pm$ 0.13                     \\
RoBERTa                            & 70.16 $\pm$ 0.19                     & 70.51 $\pm$ 0.22                     \\
Longformer$\dagger$                         & 67.31 $\pm$ 0.12                     & 67.71 $\pm$ 0.19                     \\ \hline
\end{tabular}
\caption{Summary of duplicate detection experiment results stated with 95\% confidence intervals. The $\dagger$ sign marks randomly initialized models. As we expected, our model outperformed the other tested approaches. For a broader discussion of the results and an interesting observation that the CodeBERT does not bring a statistically significant improvement over the basic RoBERTa, see  Section \ref{sec:discussion}.}
\label{tab:duplicates-results}
\end{table}

\section{Generalization to Other Tasks}\label{sec:generalization}
To explore how well our model generalizes to other tasks, we choose the \textbf{code search} task. The information retrieval seems to be close to our pre-training tasks. Therefore, we conjecture that the acquired results can be promising. For all the experiments, we use the \textit{CodeSearchNet} dataset \cite{src:code-search-net} containing approximately 2.3M examples from six different programming languages extracted from \textit{GitHub} repositories.

\subsection{Domain-Specific Pre-Training}
Since our model is pre-trained on Stack Overflow data significantly different from the \textit{CodeSearchNet} extracted from \textit{GitHub}, we employ a domain-specific pre-training to adapt our model to the target domain.

We employ the \textit{masked language modeling} (MLM) learning objective for the domain-specific pre-training. We perform 20 iterations over the \textit{CodeSearchNet} dataset following the same experimental setup as described in Section \ref{sec:pre-training-procedure}.

\subsection{Experimental Setup -- Code Search}
To fine-tune our model on the \textit{CodeSearchNet} dataset \cite{src:code-search-net}, we utilize its pre-processed version from the authors of CodeBERT \cite{src:code_bert_orig} since it comes with negative examples, unlike the original dataset distribution. In our experiments, we train a separate model for each of the six available programming languages and compare our results with the results obtained using the CodeBERT \cite{src:code_bert_orig}, RoBERTa \cite{src:roberta}, and randomly initialized Longformer \cite{src:longformer}.

For all of the experiments, we employ the \verb|AutoModelForSequenceClassification| class from the \textit{Hugging Face's Transformers} \cite{src:huggingface} library. The models for sequence classification come with an in-build classification head that operates over the pooled output of the base model.

Similarly to the duplicate detection experiments, we perform the fine-tuning on two NVidia A100 GPUs for 24 hours with a batch size of 64 examples. For optimization, we also employ the Adam \cite{src:adam} optimizer with a \textit{learning rate} of $1\mathrm{e}{-5}$. Furthermore, we utilize \textit{learning rate warmup} during the first 256 batches and apply \textit{linear learning rate decay} to zero.  

\subsection{Results}\label{sec:code-searh-results}
In the case of the code search task, we use the F1 score metric to evaluate the achieved results. The complete summary of the results with 95\% confidence intervals can be found in Table \ref{tab:search-results-1}. The results show that both the \textit{CodeBERT} \cite{src:code_bert_orig} and \textit{RoBERTa} \cite{src:roberta} significantly outperform our model in the code search task. 

\begin{table*}[ht!]
\catcode`\-=12
\begin{adjustbox}{width=\linewidth,center}
\begin{tabular}{lrrrrrr}
\hline
\textbf{Model}      & \multicolumn{1}{c}{\textbf{Go}} & \multicolumn{1}{c}{\textbf{Java}} & \multicolumn{1}{c}{\textbf{JavaScript}} & \multicolumn{1}{c}{\textbf{PHP}} & \multicolumn{1}{c}{\textbf{Python}} & \multicolumn{1}{c}{\textbf{Ruby}} \\ \hline
MQDD      & 95.33 $\pm$ 0.04      & 80.11 $\pm$ 0.15        & 70.09 $\pm$ 0.48              & 85.58 $\pm$ 0.16       & 84.14 $\pm$ 0.48          & 82.77 $\pm$ 0.31        \\
\textbf{CodeBERT}   & \textbf{96.68} $\pm$ \textbf{0.06}      & \textbf{83.75} $\pm$ \textbf{0.06}        & \textbf{83.42} $\pm$ \textbf{0.06}              & \textbf{88.50} $\pm$ \textbf{0.03}       & \textbf{88.25} $\pm$ \textbf{0.12}          & \textbf{87.22} $\pm$ \textbf{0.31}        \\
RoBERTa    & 95.94 $\pm$ 0.06      & 81.58 $\pm$ 0.23        & 80.35 $\pm$ 0.25              & 86.78 $\pm$ 0.09       & 86.02 $\pm$ 0.11          & 84.06 $\pm$ 0.20        \\
Longformer$\dagger$ & 66.62 $\pm$ 0.14      & 66.51 $\pm$ 0.24        & 66.71 $\pm$ 0.15              & 66.68 $\pm$ 0.06       & 66.71 $\pm$ 0.10          & 66.74 $\pm$ 0.15        \\ \hline
\end{tabular}
\end{adjustbox}
\caption{Results summary of \textit{code search} experiments in six different programming languages. The F1 score is stated in percents with 95\% confidence intervals. The best results in each language are highlighted in bold. The $\dagger$ sign marks randomly initialized models. The results show that the CodeBERT shows better results than our MQDD model. As discussed in Section \ref{sec:discussion}, we conjecture that this might be caused by a \textit{negative transfer} effect. Nevertheless, our model is specifically designed for duplicate question detection, and therefore, we do not place much importance on the poorer results in the code search task.}
\label{tab:search-results-1}
\end{table*}

\section{Discussion}\label{sec:discussion}
As the results stated in Sections \ref{sec:duplicates-results} and \ref{sec:code-searh-results} suggest, our model excels in detecting duplicates but lags in source code retrieval. We expected the dominance of our model in the duplication detection task. However, an interesting observation is that the pre-training of the CodeBERT initialized using the RoBERTa's weights does not bring any improvement when applied to the duplicate detection. On the other hand, it is very surprising that our MQDD model does not perform comparably well as the CodeBERT on the code search since our pre-training objectives require the model to build a deep understanding of the processed source code. 

This can be explained by the fact that the datasets used for pre-training of both models have very different characteristics. The SOD does not contain source code from a constrained set of six programming languages (see Table \ref{tab:dataset-language-analysis}), as in the case of the CodeBERT. Therefore, our model may produce representations of all programming languages in average quality. In contrast, the CodeBERT may produce high-quality representations in the six programming languages it was pre-trained on but very bad representations of the other programming languages. This would also explain why CodeBERT does not perform so well on duplicates; it excels in processing the six programming languages but fails to generalize to other abundantly contained languages in the Stack Overflow dataset.

However, the offered explanation does not cover that RoBERTa, whose pre-training dataset did not contain any source code, outperforms our model in the code search task. We speculate that this can be caused by the MQDD model being trapped in its local optimum due to its pre-training designed especially for the duplicate detection. This can make it difficult to get out of this local optimum when fine-tuned on a slightly different dataset and task. This phenomenon is often referred to as a \textit{negative transfer} \cite{src:negative-transfer-1, src:negative-transfer-domain-divergence} and can be caused, among other things, by the discrepancy between the pre-training and fine-tuning domains.

Given that our research aimed to build a model designed directly for the detection of duplicates on platforms such as Stack Overflow or Quora, it can be stated that the results we achieve are satisfactory. Our resulting model far exceeds the results achieved by competitive work on a task that can be perceived as more demanding due to the need to process a general source language and distinguish seemingly insignificant semantic nuances. For example, questions \textit{"How to implement a producer-consumer in Java"} and \textit{"How to implement a producer-consumer in C++"} must be identified as different since the answers would significantly differ.

\section{Future Work}\label{sec:future_work}

Our work opens up further opportunities to build on our current research. First of all, it would be interesting to explore methods that would eliminate the effect of negative transfer and thus allow the use of our pre-trained model in other tasks. 

Furthermore, the follow-up work can integrate our model into a production-ready duplicate detection system employing a fast vector space search library such as \textit{Faiss}.%\footnote{\url{https://engineering.fb.com/2017/03/29/data\%2Dinfrastructure/faiss\%2Da\%2Dlibrary\%2Dfor\%2Defficient\%2Dsimilarity\%2Dsearch/}}.

The proposed system can be further extended by a duplicate detection model that jointly processes both questions allowing the attention mechanism to attend across both inputs. Such a model can potentially achieve better results and be deployed along with our two-tower-based model. Our two-tower model would then be used to filter out candidate duplicate questions. Afterward, the cross-attention model could verify that the candidate questions are indeed duplicates more accurately.

\section{Conclusion}\label{sec:conclusion}

This work presents a new pre-trained BERT-like model that detects duplicate threads on programming-related discussion platforms. Based on the Longformer architecture, the presented model is pre-trained on our novel pre-training objectives (\textit{QA} and \textit{SP}) that aim to target the duplicate detection task. The comparison with the competitive CodeBERT model shows that our model outperforms other approaches, suggesting the effectiveness of our learning objectives. Furthermore, we investigated the generalization capabilities of our model by applying it to a code retrieval task. In this task, it turned out that our model does not exceed the results achieved with either CodeBERT or the more general RoBERTa model. We attribute these findings to the significant differences between our pre-training dataset and the evaluation dataset for the code search task. Therefore, we consider our model an excellent choice for solving duplicate detection. However, it seems to be too specialized to solve other tasks well.

Our models are publicly available for research purposes in our Hugging Face\footnote{\url{https://huggingface.co/UWB-AIR}} and GitHub\footnote{\url{https://github.com/kiv-air/MQDD}} repositories.

\section*{Acknowledgments}\label{sec:acknowledgements}
This work has been supported by Grant No. SGS-2022-016 Advanced methods of data processing and analysis. Computational resources were supplied by the project "e-Infrastruktura CZ" (e-INFRA CZ LM2018140 ) supported by the Ministry of Education, Youth and Sports of the Czech Republic.

\bibliography{naacl2021}

\begin{thebibliography}{49}
\expandafter\ifx\csname natexlab\endcsname\relax\def\natexlab#1{#1}\fi

\bibitem[{Arkhipov et~al.(2019)Arkhipov, Trofimova, Kuratov, and
  Sorokin}]{src:slavic-bert}
Mikhail Arkhipov, Maria Trofimova, Yuri Kuratov, and Alexey Sorokin. 2019.
\newblock \href {https://doi.org/10.18653/v1/W19-3712} {Tuning multilingual
  transformers for language-specific named entity recognition}.
\newblock In \emph{Proceedings of the 7th Workshop on Balto-Slavic Natural
  Language Processing}, pages 89--93, Florence, Italy. Association for
  Computational Linguistics.

\bibitem[{Arwan et~al.(2015)Arwan, Rochimah, and Akbar}]{src:code-search-3}
Achmad Arwan, Siti Rochimah, and Rizky~Januar Akbar. 2015.
\newblock \href {https://doi.org/10.1109/ICoICT.2015.7231439} {Source code
  retrieval on stackoverflow using lda}.
\newblock In \emph{2015 3rd International Conference on Information and
  Communication Technology (ICoICT)}, pages 295--299.

\bibitem[{Bahdanau et~al.(2014)Bahdanau, Cho, and
  Bengio}]{src:attention-mechanism}
Dzmitry Bahdanau, Kyunghyun Cho, and Y.~Bengio. 2014.
\newblock Neural machine translation by jointly learning to align and
  translate.
\newblock \emph{ArXiv}, 1409.

\bibitem[{Balog et~al.(2017)Balog, Gaunt, Brockschmidt, Nowozin, and
  Tarlow}]{balog2017deepcoder}
M~Balog, AL~Gaunt, M~Brockschmidt, S~Nowozin, and D~Tarlow. 2017.
\newblock Deepcoder: Learning to write programs.
\newblock In \emph{International Conference on Learning Representations (ICLR
  2017)}. OpenReview. net.

\bibitem[{Beltagy et~al.(2020)Beltagy, Peters, and Cohan}]{src:longformer}
Iz~Beltagy, Matthew~E. Peters, and Arman Cohan. 2020.
\newblock \href {http://arxiv.org/abs/2004.05150} {Longformer: The
  long-document transformer}.
\newblock \emph{CoRR}, abs/2004.05150.

\bibitem[{Biewald(2020)}]{src:wandb}
Lukas Biewald. 2020.
\newblock \href {https://www.wandb.com/} {Experiment tracking with weights and
  biases}.
\newblock Software available from wandb.com.

\bibitem[{Bridle(1990)}]{src:softmax}
John~S. Bridle. 1990.
\newblock Probabilistic interpretation of feedforward classification network
  outputs, with relationships to statistical pattern recognition.
\newblock In \emph{Neurocomputing}, pages 227--236, Berlin, Heidelberg.
  Springer Berlin Heidelberg.

\bibitem[{Brown et~al.(2020)Brown, Mann, Ryder, Subbiah, Kaplan, Dhariwal,
  Neelakantan, Shyam, Sastry, Askell, Agarwal, Herbert{-}Voss, Krueger,
  Henighan, Child, Ramesh, Ziegler, Wu, Winter, Hesse, Chen, Sigler, Litwin,
  Gray, Chess, Clark, Berner, McCandlish, Radford, Sutskever, and
  Amodei}]{src:GPT}
Tom~B. Brown, Benjamin Mann, Nick Ryder, Melanie Subbiah, Jared Kaplan,
  Prafulla Dhariwal, Arvind Neelakantan, Pranav Shyam, Girish Sastry, Amanda
  Askell, Sandhini Agarwal, Ariel Herbert{-}Voss, Gretchen Krueger, Tom
  Henighan, Rewon Child, Aditya Ramesh, Daniel~M. Ziegler, Jeffrey Wu, Clemens
  Winter, Christopher Hesse, Mark Chen, Eric Sigler, Mateusz Litwin, Scott
  Gray, Benjamin Chess, Jack Clark, Christopher Berner, Sam McCandlish, Alec
  Radford, Ilya Sutskever, and Dario Amodei. 2020.
\newblock \href {http://arxiv.org/abs/2005.14165} {Language models are few-shot
  learners}.
\newblock \emph{CoRR}, abs/2005.14165.

\bibitem[{Cabrera~Lozoya et~al.(2021)Cabrera~Lozoya, Baumann, Sabetta, and
  Bezzi}]{cabrera2021commit2vec}
Roc{\'\i}o Cabrera~Lozoya, Arnaud Baumann, Antonino Sabetta, and Michele Bezzi.
  2021.
\newblock Commit2vec: Learning distributed representations of code changes.
\newblock \emph{SN Computer Science}, 2(3):1--16.

\bibitem[{Chen et~al.(2021)Chen, Tworek, Jun, Yuan, de~Oliveira~Pinto, Kaplan,
  Edwards, Burda, Joseph, Brockman, Ray, Puri, Krueger, Petrov, Khlaaf, Sastry,
  Mishkin, Chan, Gray, Ryder, Pavlov, Power, Kaiser, Bavarian, Winter, Tillet,
  Such, Cummings, Plappert, Chantzis, Barnes, Herbert{-}Voss, Guss, Nichol,
  Paino, Tezak, Tang, Babuschkin, Balaji, Jain, Saunders, Hesse, Carr, Leike,
  Achiam, Misra, Morikawa, Radford, Knight, Brundage, Murati, Mayer, Welinder,
  McGrew, Amodei, McCandlish, Sutskever, and Zaremba}]{src:codex}
Mark Chen, Jerry Tworek, Heewoo Jun, Qiming Yuan, Henrique~Ponde
  de~Oliveira~Pinto, Jared Kaplan, Harrison Edwards, Yuri Burda, Nicholas
  Joseph, Greg Brockman, Alex Ray, Raul Puri, Gretchen Krueger, Michael Petrov,
  Heidy Khlaaf, Girish Sastry, Pamela Mishkin, Brooke Chan, Scott Gray, Nick
  Ryder, Mikhail Pavlov, Alethea Power, Lukasz Kaiser, Mohammad Bavarian,
  Clemens Winter, Philippe Tillet, Felipe~Petroski Such, Dave Cummings,
  Matthias Plappert, Fotios Chantzis, Elizabeth Barnes, Ariel Herbert{-}Voss,
  William~Hebgen Guss, Alex Nichol, Alex Paino, Nikolas Tezak, Jie Tang, Igor
  Babuschkin, Suchir Balaji, Shantanu Jain, William Saunders, Christopher
  Hesse, Andrew~N. Carr, Jan Leike, Joshua Achiam, Vedant Misra, Evan Morikawa,
  Alec Radford, Matthew Knight, Miles Brundage, Mira Murati, Katie Mayer, Peter
  Welinder, Bob McGrew, Dario Amodei, Sam McCandlish, Ilya Sutskever, and
  Wojciech Zaremba. 2021.
\newblock \href {http://arxiv.org/abs/2107.03374} {Evaluating large language
  models trained on code}.
\newblock \emph{CoRR}, abs/2107.03374.

\bibitem[{Chen and Monperrus(2019)}]{src:source_code_embedding_literature}
Zimin Chen and Martin Monperrus. 2019.
\newblock \href {http://arxiv.org/abs/1904.03061} {A literature study of
  embeddings on source code}.
\newblock \emph{CoRR}, abs/1904.03061.

\bibitem[{Clark et~al.(2020)Clark, Luong, Le, and Manning}]{src:electra}
Kevin Clark, Minh{-}Thang Luong, Quoc~V. Le, and Christopher~D. Manning. 2020.
\newblock \href {http://arxiv.org/abs/2003.10555} {{ELECTRA:} pre-training text
  encoders as discriminators rather than generators}.
\newblock \emph{CoRR}, abs/2003.10555.

\bibitem[{Deng et~al.(2020)Deng, Awadallah, Meek, Polozov, Sun, and
  Richardson}]{deng2020structure}
Xiang Deng, Ahmed~Hassan Awadallah, Christopher Meek, Oleksandr Polozov, Huan
  Sun, and Matthew Richardson. 2020.
\newblock Structure-grounded pretraining for text-to-sql.
\newblock \emph{arXiv preprint arXiv:2010.12773}.

\bibitem[{Devlin et~al.(2018)Devlin, Chang, Lee, and Toutanova}]{src:bert}
Jacob Devlin, Ming{-}Wei Chang, Kenton Lee, and Kristina Toutanova. 2018.
\newblock \href {http://arxiv.org/abs/1810.04805} {{BERT:} pre-training of deep
  bidirectional transformers for language understanding}.
\newblock \emph{CoRR}, abs/1810.04805.

\bibitem[{Feng et~al.(2020)Feng, Guo, Tang, Duan, Feng, Gong, Shou, Qin, Liu,
  Jiang, and Zhou}]{src:code_bert_orig}
Zhangyin Feng, Daya Guo, Duyu Tang, Nan Duan, Xiaocheng Feng, Ming Gong, Linjun
  Shou, Bing Qin, Ting Liu, Daxin Jiang, and Ming Zhou. 2020.
\newblock \href {http://arxiv.org/abs/2002.08155} {Codebert: {A} pre-trained
  model for programming and natural languages}.
\newblock \emph{CoRR}, abs/2002.08155.

\bibitem[{Gupta and Sundaresan(2018)}]{gupta2018intelligent}
Anshul Gupta and Neel Sundaresan. 2018.
\newblock Intelligent code reviews using deep learning.
\newblock In \emph{Proceedings of the 24th ACM SIGKDD International Conference
  on Knowledge Discovery and Data Mining (KDD’18) Deep Learning Day}.

\bibitem[{Heyman and Cutsem(2020)}]{src:code-search-1}
Geert Heyman and Tom~Van Cutsem. 2020.
\newblock \href {http://arxiv.org/abs/2008.12193} {Neural code search
  revisited: Enhancing code snippet retrieval through natural language intent}.
\newblock \emph{CoRR}, abs/2008.12193.

\bibitem[{Hindle et~al.(2012)Hindle, Barr, Su, Gabel, and
  Devanbu}]{src:on-naturalness}
Abram Hindle, Earl Barr, Zhendong Su, Mark Gabel, and Premkumar Devanbu. 2012.
\newblock \href {https://doi.org/10.1109/ICSE.2012.6227135} {On the naturalness
  of software}.
\newblock \emph{Proceedings - International Conference on Software
  Engineering}, pages 837--847.

\bibitem[{Husain et~al.(2019)Husain, Wu, Gazit, Allamanis, and
  Brockschmidt}]{src:code-search-net}
Hamel Husain, Ho{-}Hsiang Wu, Tiferet Gazit, Miltiadis Allamanis, and Marc
  Brockschmidt. 2019.
\newblock \href {http://arxiv.org/abs/1909.09436} {Codesearchnet challenge:
  Evaluating the state of semantic code search}.
\newblock \emph{CoRR}, abs/1909.09436.

\bibitem[{Johnson et~al.(2019)Johnson, Douze, and J{\'e}gou}]{src:faiss}
Jeff Johnson, Matthijs Douze, and Herv{\'e} J{\'e}gou. 2019.
\newblock Billion-scale similarity search with {GPUs}.
\newblock \emph{IEEE Transactions on Big Data}, 7(3):535--547.

\bibitem[{Kanade et~al.(2020{\natexlab{a}})Kanade, Maniatis, Balakrishnan, and
  Shi}]{src:cubert}
Aditya Kanade, Petros Maniatis, Gogul Balakrishnan, and Kensen Shi.
  2020{\natexlab{a}}.
\newblock \href {http://arxiv.org/abs/2001.00059} {Pre-trained contextual
  embedding of source code}.
\newblock \emph{CoRR}, abs/2001.00059.

\bibitem[{Kanade et~al.(2020{\natexlab{b}})Kanade, Maniatis, Balakrishnan, and
  Shi}]{src:LSTM}
Aditya Kanade, Petros Maniatis, Gogul Balakrishnan, and Kensen Shi.
  2020{\natexlab{b}}.
\newblock \href {http://arxiv.org/abs/2001.00059} {Pre-trained contextual
  embedding of source code}.
\newblock \emph{CoRR}, abs/2001.00059.

\bibitem[{Kingma and Ba(2014)}]{src:adam}
Diederik Kingma and Jimmy Ba. 2014.
\newblock \href {https://arxiv.org/abs/1412.6980} {Adam: A method for
  stochastic optimization}.
\newblock \emph{International Conference on Learning Representations}.

\bibitem[{Kitaev et~al.(2020)Kitaev, Kaiser, and Levskaya}]{src:reformer}
Nikita Kitaev, Lukasz Kaiser, and Anselm Levskaya. 2020.
\newblock \href {http://arxiv.org/abs/2001.04451} {Reformer: The efficient
  transformer}.
\newblock \emph{CoRR}, abs/2001.04451.

\bibitem[{Le et~al.(2020)Le, Chen, and Babar}]{src:deep-learning-source-code}
Triet Huynh~Minh Le, Hao Chen, and Muhammad~Ali Babar. 2020.
\newblock \href {http://arxiv.org/abs/2002.05442} {Deep learning for source
  code modeling and generation: Models, applications and challenges}.
\newblock \emph{CoRR}, abs/2002.05442.

\bibitem[{Liu et~al.(2020)Liu, Gao, Chen, Yiu, and Liu}]{liu2020atom}
Shangqing Liu, Cuiyun Gao, Sen Chen, Nie~Lun Yiu, and Yang Liu. 2020.
\newblock Atom: Commit message generation based on abstract syntax tree and
  hybrid ranking.
\newblock \emph{IEEE Transactions on Software Engineering}.

\bibitem[{Liu et~al.(2019)Liu, Ott, Goyal, Du, Joshi, Chen, Levy, Lewis,
  Zettlemoyer, and Stoyanov}]{src:roberta}
Yinhan Liu, Myle Ott, Naman Goyal, Jingfei Du, Mandar Joshi, Danqi Chen, Omer
  Levy, Mike Lewis, Luke Zettlemoyer, and Veselin Stoyanov. 2019.
\newblock \href {http://arxiv.org/abs/1907.11692} {Roberta: {A} robustly
  optimized {BERT} pretraining approach}.
\newblock \emph{CoRR}, abs/1907.11692.

\bibitem[{Liu et~al.(2021)Liu, Jiang, Hu, Shi, and Fung}]{src:bert-ner}
Zihan Liu, Feijun Jiang, Yuxiang Hu, Chen Shi, and Pascale Fung. 2021.
\newblock \href {http://arxiv.org/abs/2112.00405} {{NER-BERT:} {A} pre-trained
  model for low-resource entity tagging}.
\newblock \emph{CoRR}, abs/2112.00405.

\bibitem[{Lu et~al.(2019)Lu, Batra, Parikh, and Lee}]{src:vilbert}
Jiasen Lu, Dhruv Batra, Devi Parikh, and Stefan Lee. 2019.
\newblock \href {http://arxiv.org/abs/1908.02265} {Vilbert: Pretraining
  task-agnostic visiolinguistic representations for vision-and-language tasks}.
\newblock \emph{CoRR}, abs/1908.02265.

\bibitem[{McCann et~al.(2017)McCann, Bradbury, Xiong, and
  Socher}]{src:contextual-2}
Bryan McCann, James Bradbury, Caiming Xiong, and Richard Socher. 2017.
\newblock \href {http://arxiv.org/abs/1708.00107} {Learned in translation:
  Contextualized word vectors}.
\newblock \emph{CoRR}, abs/1708.00107.

\bibitem[{Nair and Hinton(2010)}]{src:relu}
Vinod Nair and Geoffrey~E. Hinton. 2010.
\newblock Rectified linear units improve restricted boltzmann machines.
\newblock In \emph{Proceedings of the 27th International Conference on
  International Conference on Machine Learning}, ICML'10, page 807–814,
  Madison, WI, USA. Omnipress.

\bibitem[{Peters et~al.(2018)Peters, Neumann, Iyyer, Gardner, Clark, Lee, and
  Zettlemoyer}]{src:contextual-1}
Matthew~E. Peters, Mark Neumann, Mohit Iyyer, Matt Gardner, Christopher Clark,
  Kenton Lee, and Luke Zettlemoyer. 2018.
\newblock \href {https://doi.org/10.18653/v1/N18-1202} {Deep contextualized
  word representations}.
\newblock In \emph{Proceedings of the 2018 Conference of the North {A}merican
  Chapter of the Association for Computational Linguistics: Human Language
  Technologies, Volume 1 (Long Papers)}, pages 2227--2237, New Orleans,
  Louisiana. Association for Computational Linguistics.

\bibitem[{Raffel et~al.(2019)Raffel, Shazeer, Roberts, Lee, Narang, Matena,
  Zhou, Li, and Liu}]{src:T5}
Colin Raffel, Noam Shazeer, Adam Roberts, Katherine Lee, Sharan Narang, Michael
  Matena, Yanqi Zhou, Wei Li, and Peter~J. Liu. 2019.
\newblock \href {http://arxiv.org/abs/1910.10683} {Exploring the limits of
  transfer learning with a unified text-to-text transformer}.
\newblock \emph{CoRR}, abs/1910.10683.

\bibitem[{Reimers and Gurevych(2019)}]{src:bert-textual-similarity}
Nils Reimers and Iryna Gurevych. 2019.
\newblock \href {http://arxiv.org/abs/1908.10084} {Sentence-bert: Sentence
  embeddings using siamese bert-networks}.
\newblock \emph{CoRR}, abs/1908.10084.

\bibitem[{Rosenstein et~al.(2005)Rosenstein, Marx, Kaelbling, and
  Dietterich}]{src:negative-transfer-1}
Michael Rosenstein, Zvika Marx, Leslie Kaelbling, and Thomas Dietterich. 2005.
\newblock To transfer or not to transfer.

\bibitem[{Sachdev et~al.(2018)Sachdev, Li, Luan, Kim, Sen, and
  Chandra}]{src:code-search-2}
Saksham Sachdev, Hongyu Li, Sifei Luan, Seohyun Kim, Koushik Sen, and Satish
  Chandra. 2018.
\newblock \href {https://doi.org/10.1145/3211346.3211353} {Retrieval on source
  code: A neural code search}.
\newblock In \emph{Proceedings of the 2nd ACM SIGPLAN International Workshop on
  Machine Learning and Programming Languages}, MAPL 2018, page 31–41, New
  York, NY, USA. Association for Computing Machinery.

\bibitem[{Schuster and Paliwal(1997)}]{src:bi-rnn}
M.~Schuster and K.K. Paliwal. 1997.
\newblock \href {https://doi.org/10.1109/78.650093} {Bidirectional recurrent
  neural networks}.
\newblock \emph{IEEE Transactions on Signal Processing}, 45(11):2673--2681.

\bibitem[{Schuster and Nakajima(2012)}]{src:word-piece}
Mike Schuster and Kaisuke Nakajima. 2012.
\newblock \href {https://doi.org/10.1109/ICASSP.2012.6289079} {Japanese and
  korean voice search}.
\newblock In \emph{2012 IEEE International Conference on Acoustics, Speech and
  Signal Processing (ICASSP)}, pages 5149--5152.

\bibitem[{Sun et~al.(2019{\natexlab{a}})Sun, Myers, Vondrick, Murphy, and
  Schmid}]{src:videobert}
Chen Sun, Austin Myers, Carl Vondrick, Kevin Murphy, and Cordelia Schmid.
  2019{\natexlab{a}}.
\newblock \href {http://arxiv.org/abs/1904.01766} {Videobert: {A} joint model
  for video and language representation learning}.
\newblock \emph{CoRR}, abs/1904.01766.

\bibitem[{Sun et~al.(2019{\natexlab{b}})Sun, Huang, and
  Qiu}]{src:bert-sentiment}
Chi Sun, Luyao Huang, and Xipeng Qiu. 2019{\natexlab{b}}.
\newblock \href {http://arxiv.org/abs/1903.09588} {Utilizing {BERT} for
  aspect-based sentiment analysis via constructing auxiliary sentence}.
\newblock \emph{CoRR}, abs/1903.09588.

\bibitem[{Sun et~al.(2022)Sun, Fang, Chen, Tao, Han, and Zhang}]{sun2022code}
Weisong Sun, Chunrong Fang, Yuchen Chen, Guanhong Tao, Tingxu Han, and Quanjun
  Zhang. 2022.
\newblock Code search based on context-aware code translation.
\newblock \emph{arXiv preprint arXiv:2202.08029}.

\bibitem[{Tufano et~al.(2018)Tufano, Watson, Bavota, Di~Penta, White, and
  Poshyvanyk}]{tufano2018deep}
Michele Tufano, Cody Watson, Gabriele Bavota, Massimiliano Di~Penta, Martin
  White, and Denys Poshyvanyk. 2018.
\newblock Deep learning similarities from different representations of source
  code.
\newblock In \emph{2018 IEEE/ACM 15th International Conference on Mining
  Software Repositories (MSR)}, pages 542--553. IEEE.

\bibitem[{Ullah et~al.(2019)Ullah, Wang, Jabbar, Al-Turjman, and
  Alazab}]{ullah2019source}
Farhan Ullah, Junfeng Wang, Sohail Jabbar, Fadi Al-Turjman, and Mamoun Alazab.
  2019.
\newblock Source code authorship attribution using hybrid approach of program
  dependence graph and deep learning model.
\newblock \emph{IEEE Access}, 7:141987--141999.

\bibitem[{Vaswani et~al.(2017)Vaswani, Shazeer, Parmar, Uszkoreit, Jones,
  Gomez, Kaiser, and Polosukhin}]{src:transformer}
Ashish Vaswani, Noam Shazeer, Niki Parmar, Jakob Uszkoreit, Llion Jones,
  Aidan~N. Gomez, Lukasz Kaiser, and Illia Polosukhin. 2017.
\newblock \href {http://arxiv.org/abs/1706.03762} {Attention is all you need}.
\newblock \emph{CoRR}, abs/1706.03762.

\bibitem[{Wang et~al.(2020)Wang, Zhang, and
  Jiang}]{src:duplicate-detection-lstm}
Liting Wang, Li~Zhang, and Jing Jiang. 2020.
\newblock \href {https://doi.org/10.1109/ACCESS.2020.2968391} {Duplicate
  question detection with deep learning in stack overflow}.
\newblock \emph{IEEE Access}, 8:25964--25975.

\bibitem[{Wang et~al.(2021)Wang, Wang, Joty, and Hoi}]{src:code-t5}
Yue Wang, Weishi Wang, Shafiq~R. Joty, and Steven C.~H. Hoi. 2021.
\newblock \href {http://arxiv.org/abs/2109.00859} {Codet5: Identifier-aware
  unified pre-trained encoder-decoder models for code understanding and
  generation}.
\newblock \emph{CoRR}, abs/2109.00859.

\bibitem[{Wolf et~al.(2020)Wolf, Debut, Sanh, Chaumond, Delangue, Moi, Cistac,
  Rault, Louf, Funtowicz, Davison, Shleifer, von Platen, Ma, Jernite, Plu, Xu,
  Scao, Gugger, Drame, Lhoest, and Rush}]{src:huggingface}
Thomas Wolf, Lysandre Debut, Victor Sanh, Julien Chaumond, Clement Delangue,
  Anthony Moi, Pierric Cistac, Tim Rault, Rémi Louf, Morgan Funtowicz, Joe
  Davison, Sam Shleifer, Patrick von Platen, Clara Ma, Yacine Jernite, Julien
  Plu, Canwen Xu, Teven~Le Scao, Sylvain Gugger, Mariama Drame, Quentin Lhoest,
  and Alexander~M. Rush. 2020.
\newblock \href {https://www.aclweb.org/anthology/2020.emnlp-demos.6}
  {Transformers: State-of-the-art natural language processing}.
\newblock In \emph{Proceedings of the 2020 Conference on Empirical Methods in
  Natural Language Processing: System Demonstrations}, pages 38--45, Online.
  Association for Computational Linguistics.

\bibitem[{Yin and Neubig(2017)}]{yin2017syntactic}
Pengcheng Yin and Graham Neubig. 2017.
\newblock A syntactic neural model for general-purpose code generation.
\newblock In \emph{Proceedings of the 55th Annual Meeting of the Association
  for Computational Linguistics (Volume 1: Long Papers)}, pages 440--450.

\bibitem[{Zhang et~al.(2020)Zhang, Deng, and
  Wu}]{src:negative-transfer-domain-divergence}
Wen Zhang, Lingfei Deng, and Dongrui Wu. 2020.
\newblock \href {http://arxiv.org/abs/2009.00909} {A survey on negative
  transfer}.
\newblock \emph{CoRR}, abs/2009.00909.

\end{thebibliography}
\bibliographystyle{acl_natbib}

\clearpage
\appendix
\section{Dataset Pre-processing}\label{sec:dataset-preprocessing}
The data retrieved from the Stack Overflow data dump contain an HTML markup that needs to be pre-processed before being used to train a neural network. Furthermore, the natural language and source code snippets are mixed in a single HTML document, so we need to separate those two parts. 

We use the \verb|BeautifulSoup|\footnote{\url{https://www.crummy.com/software/BeautifulSoup/bs4/doc/}} library to extract the textual data from the HTML markup. To do so, we remove all content enclosed in \verb|<code></code>| tags and strip all the remaining HTML tags. Afterward, we remove all newline characters and multiple subsequent space characters induced by stripping the HTML tags.

On the other hand, while pre-processing the code snippets, we first extract all content from \verb|<pre><code></code></pre>| using the \verb|BeautifulSoup| library and throw away the rest. Afterward, we remove the newlines and multiple spaces, as in the case of the textual part. 

\section{Longformer Model Configuration}\label{sec:longformer-configuration}

The implementation of the Longformer model that we employ in the pre-training is the \verb|transformers.LongformerModel|\footnote{\url{https://huggingface.co/docs/transformers/model_doc/longformer\#transformers.LongformerModel}} from \textit{HuggingFace Transformers} library. Below, we provide a detailed listing of the model's parameters.

\begin{itemize}
\itemsep0em 
    \item attention\_probs\_dropout\_prob = 0.1
	\item attention\_window = 256
	\item hidden\_act = gelu
	\item hidden\_dropout\_prob = 0.1
	\item hidden\_size = 768
	\item initializer\_range = 0.02
	\item intermediate\_size = 3072
	\item layer\_norm\_eps = 1e-12
	\item max\_position\_embeddings = 1026
	\item num\_attention\_heads = 12
	\item num\_hidden\_layers = 12
	\item position\_embedding\_type = absolute
	\item vocab\_size = 50256
	\item intermediate\_layer\_dim ($D$) = 1000
\end{itemize}

\section{Duplicate Detection Hyperparameters}\label{sec:duplicates-hyperparameters}
For fine-tuning our MQDD model on the duplicate detection task, we employ the Adam optimizer with an initial learning rate of $6.35\mathrm{e}{-6}$. We train the model on sequences of 256 subword tokens with a batch size of 100 examples. Additionally, we use an L2 normalization with a normalization factor set to 0.043. Another regularization method we employ is the dropout with the following configuration:

\begin{itemize}
    \itemsep0em 
    \item attention dropout in the Longformer = 0.2
    \item hidden dropout in the Longformer = 0.5
    \item dropout at the first linear layer of the classification head = 0.26
    \item dropout at the second linear layer of the classification head = 0.2
\end{itemize}

\section{Stack Overflow Dataset Structure}\label{sec:sod-structure}
The \textit{Stack Overflow Dataset} (SOD) consists of a metadata file and several data files. Each line of the metadata file (\verb|dataset_meta.csv|) contains a \textit{JSON} array with the following information:

\begin{itemize}
    \itemsep0em 
    \item \textbf{question\_id} - identifier of the question in format \verb|<id>-<page>| (in our case the \verb|page = stackoverflow|)
    \item \textbf{answer\_id} - identifier of the answer in format \verb|<id>-<page>| (in our case the \verb|page = stackoverflow|)
    \item \textbf{title} - title of the question
    \item \textbf{tags} - tags associated with the question
    \item \textbf{is\_accepted} - boolean flag indicating whether the answer represents an accepted answer for the question
\end{itemize}

The dataset export is organized in such a way that $i$-th row in the metadata file corresponds to training examples located on the $i$-th row in the data files. There are six different data file types, each comprising training examples of different \textit{input pair types} (described in Section \ref{sec:pretrainig-objectives}). A complete list of the data file types follows:

\begin{itemize}
    \itemsep0em 
    \item \verb|dataset_AC_AT.csv| - code from an answer with text from the same answer
    \item \verb|dataset_QC_AC.csv| - code from a question with code from a related answer 
    \item \verb|dataset_QC_AT.csv| - code from a question with text from a related answer
    \item \verb|dataset_QC_QT.csv| - code from a question with text from the same question
    \item \verb|dataset_QT_AC.csv| - text from a question with code from a related answer
    \item \verb|dataset_QT_AT.csv| - text from a question with text from a related answer
\end{itemize}

Each row in the data file then represents a single example whose metadata can be obtained from a corresponding row in the metadata file. A training example is represented by a \textit{JSON} array containing two strings. For example, in the \verb|dataset_QC_AC.csv|, the first element in the array contains code from a question, whereas the second element contains code from the related answer. It shall be noted that the dataset export does not contain negative examples since they would significantly increase the disk space required for storing the dataset. The negative examples must be randomly sampled during pre-processing, as discussed in Section \ref{sec:pretraining-dataset}.

Since the resulting dataset takes up a lot of disk space, we split the individual data files and the metadata file into nine smaller ones. Therefore, files such as, for example, \verb|dataset_meta_1.csv| and corresponding \verb|dataset_QC_AT_1.csv| can then be found in the repository.

\section{Stack Overflow Duplicity Dataset Structure}\label{sec:sodd-structure}
The published \textit{SODD} dataset is split into train/dev/test splits and is stored in \textit{parquet}\footnote{\url{https://parquet.apache.org/documentation/latest/}} files compressed using gzip. The data can be loaded using the \textit{pandas}\footnote{\url{https://pandas.pydata.org}} library using the following code snippet:

\begin{lstlisting}[language=Python]
!pip3 install pandas pyarrow

import pandas as pd

d=pd.read_parquet('<file>.parquet.gzip')
\end{lstlisting}

The dataframe loaded using the snippet above contains the following columns:

\begin{itemize}
    \itemsep0em 
    \item \textbf{first\_post} - HTML formatted data of the first question (contains both text and code snippets)
    \item \textbf{second\_post} - HTML formatted data of the second question (contains both text and code snippets)
    \item \textbf{first\_author} - username of the first question's author
    \item \textbf{second\_author} - username of the second question's author
    \item \textbf{label} - label determining the relationship of the two questions
    \begin{enumerate}
        \itemsep0em
        \setcounter{enumi}{-1}
        \item duplicates
        \item similar based on full-text search
        \item similar based on tags
        \item different
        \item accepted answer
    \end{enumerate}
    \item \textbf{page} - Stack Exchange page from which the questions originate (always set to \verb|stackoverflow|)
\end{itemize}

As one can see, our dataset contains accepted answers as well. Although we are not using them in our work, we included them in the dataset to open up other possibilities of using our dataset.

\end{document}